\documentclass[lettersize,journal]{IEEEtran}
\usepackage{amsmath,amsfonts}
\usepackage{algorithmic}
\usepackage{array}
\usepackage[caption=false,font=normalsize,labelfont=sf,textfont=sf]{subfig}
\usepackage{textcomp}
\usepackage{stfloats}
\usepackage{url}
\usepackage{verbatim}
\usepackage{graphicx}
\usepackage{soul}
\def\BibTeX{{\rm B\kern-.05em{\sc i\kern-.025em b}\kern-.08em
    T\kern-.1667em\lower.7ex\hbox{E}\kern-.125emX}}
\usepackage{balance}

\usepackage{booktabs}
\usepackage{amssymb}

\usepackage[pagebackref=true,breaklinks=true,letterpaper=true,colorlinks,bookmarks=false]{hyperref}

\usepackage[table,xcdraw]{xcolor}
\usepackage{colortbl}

\definecolor{baselinecolor}{gray}{.9}
\newcommand{\baseline}[1]{\cellcolor{baselinecolor}{#1}}
\newcommand{\tablestyle}[2]{\setlength{\tabcolsep}{#1}\renewcommand{\arraystretch}{#2}\centering\footnotesize}

\usepackage[capitalize]{cleveref}

\usepackage[font=small,labelfont=bf]{caption}
\newcommand{\revise}[1]{#1}
\newcommand{\red}[1]{#1}

\begin{document}

\title{Forecast-PEFT: Parameter-Efficient Fine-Tuning for Pre-trained Motion Forecasting Models}

\author{Jifeng Wang, Kaouther Messaoud, Yuejiang Liu
\\
Juergen Gall, \IEEEmembership{Member, IEEE}, Alexandre Alahi, \IEEEmembership{Member, IEEE}
\thanks{Corresponding author: Jifeng Wang.}
\thanks{J. Wang and J. Gall are with the University of Bonn, Bonn, Germany (e-mail: jfwang.cs@gmail.com; gall@iai.uni-bonn.de). J. Gall is with the Lamarr Institute for Machine Learning and Artificial Intelligence, Germany.
Y. Liu is with Stanford University, Stanford, USA (e-mail: yuejiang.liu@stanford.edu). 
K. Messaoud and A. Alahi are with the VITA Laboratory, EPFL, Lausanne, Switzerland (e-mail: kaouther.messaoudbenamor@epfl.ch; alexandre.alahi@epfl.ch).}
\thanks{\copyright~2026 IEEE. Personal use of this material is permitted. Permission from IEEE must be obtained for all other uses, in any current or future media, including reprinting/republishing this material for advertising or promotional purposes, creating new collective works, for resale or redistribution to servers or lists, or reuse of any copyrighted component of this work in other works.}
}

\markboth{ IEEE TRANSACTIONS ON INTELLIGENT TRANSPORTATION SYSTEMS}%
{Forecast-PEFT: Parameter-Efficient Fine-Tuning for Pre-trained Motion Forecasting Models}

\maketitle

\begin{abstract}
  Recent advancements in motion forecasting have been significantly accelerated by self-supervised pre-training techniques. However, current methods typically rely on extensive fine-tuning of pre-trained models for downstream tasks like trajectory prediction, a process that can be inefficient. This inefficiency primarily arises because traditional fine-tuning strategies fail to fully exploit the close alignment between motion prediction tasks and masked reconstruction objectives used during pre-training. To overcome these limitations, we introduce \textbf{Forecast-PEFT}, a parameter-efficient fine-tuning framework specifically designed for motion forecasting. Forecast-PEFT preserves most of the pre-trained model's parameters, introducing minimal yet effective adjustments through novel prompts and adapters. By doing so, Forecast-PEFT significantly reduces the number of parameters needing retraining while preserving essential pre-trained knowledge. Our method demonstrates robust generalization across multiple datasets, achieving higher accuracy with only about 17\% of the parameters required by traditional full fine-tuning approaches. Additionally, our fully fine-tuned variant, \textbf{Forecast-FT}, further enhances prediction performance, achieving improvements of up to 9.6\% over conventional baseline methods. These findings highlight Forecast-PEFT as a scalable, efficient, and effective alternative for fine-tuning pre-trained motion forecasting models.
  Code is available at \url{https://github.com/csjfwang/Forecast-PEFT}. 
\end{abstract}

\begin{IEEEkeywords}
Motion Forecasting, Trajectory Prediction, Autonomous Driving, Parameter-efficient Fine-tuning
\end{IEEEkeywords}

\revise{\section{Introduction}}
\label{sec:intro}

Trajectory prediction is a critical task in autonomous driving, essential for ensuring safety and operational efficiency \cite{ghorai2022state, jeong2019target, karle2022scenario, gomez2023efficient}. Its complexity arises from dynamic interactions among road users and varying environmental conditions. Recent advancements in self-supervised learning (SSL) have significantly improved trajectory prediction performance, with pre-trained models such as Traj-MAE \cite{chen2023traj}, SEPT \cite{lan2023sept}, and Forecast-MAE \cite{cheng2023forecast} demonstrating strong predictive capabilities.
\begin{figure*}[t]
\begin{center}
\includegraphics[width=1.0\linewidth]{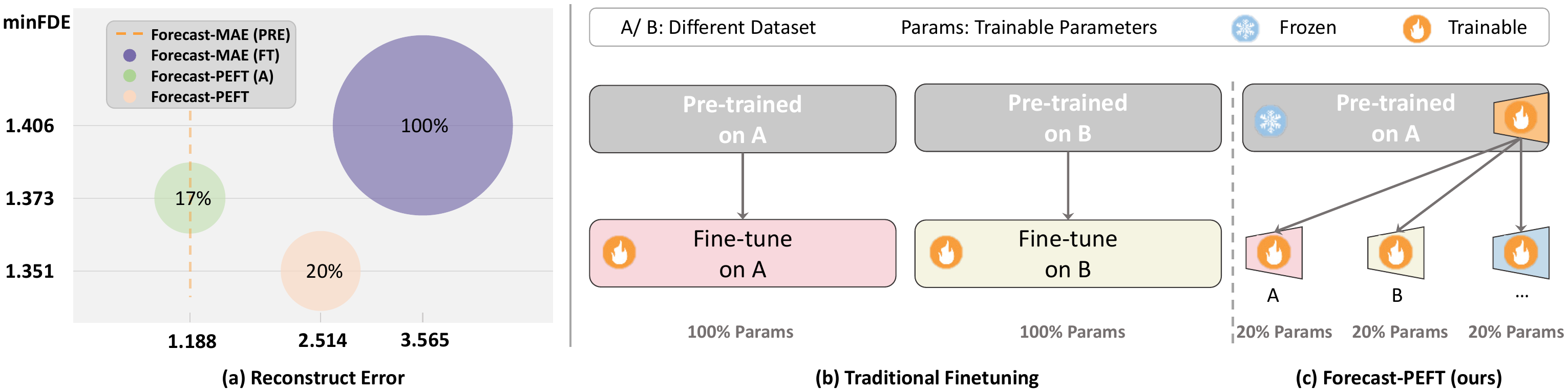}
\end{center}
   \caption{(a) Comparison of trainable parameters (percentage), performance (minFDE), and catastrophic forgetting (Reconstruction Error). The dashed line indicates the reconstruction error of the pre-trained baseline Forecast-MAE (PRE). \revise{The colored circles represent the full fine-tuned baseline Forecast-MAE (FT), and our presented PEFT methods, namely Forecast-PEFT (A), which uses only additive modules, and the default model Forecast-PEFT.}
   (b) Traditional pretraining and fine-tuning on different datasets. 
   (c) Our Forecast-PEFT method offers flexible cross-dataset fine-tuning, requiring just a single pre-training on a large dataset followed by efficient fine-tuning across various datasets. \revise{Its trainable parameters, acting as a plug-in module, require only 20\% of the parameters for adaptation to each new dataset.}
}
\label{fig:start}
\end{figure*}
However, current SSL approaches commonly utilize only pre-trained encoders and rely on extensive fine-tuning of randomly initialized decoders for downstream tasks. This practice underutilizes the potential of pre-trained decoders, especially given that motion forecasting inherently aligns closely with pre-training tasks such as trajectory reconstruction. Moreover, introducing a newly initialized decoder during fine-tuning can adversely affect the encoder's learned representations, causing catastrophic forgetting and diminishing the benefits of pre-training.

\revise{
In addition, traditional fine-tuning approaches require maintaining separate weight sets for each dataset, which is impractical in real-world scenarios due to computational and storage overheads. This limitation is particularly critical for motion forecasting, where models are expected to generalize across diverse datasets and deployment conditions.

To address these challenges, we propose Forecast-PEFT, a parameter-efficient fine-tuning framework tailored to motion forecasting. Forecast-PEFT retains both the encoder and decoder from pre-training and introduces lightweight modules to enable efficient adaptation: Contextual Embedding Prompts (CEPs) bridge the gap between masked inputs in pre-training and history-only inputs during fine-tuning; Modality-Control Prompts (MCPs) guide the decoder from masked reconstruction to diverse, multi-modal future prediction; and Parallel Adapters (PAs) capture fine-grained local trajectory cues without extensive parameter updates. These modules complement each other across encoder inputs, decoder outputs, and intermediate representations, enabling effective adaptation with minimal trainable parameters.

This design improves fine-tuning efficiency within a single dataset and naturally supports cross-dataset adaptation. In practice, our framework follows a two-stage process: an initial pre-training phase to learn universal trajectory representations, followed by parameter-efficient specialization via Forecast-PEFT modules. This workflow reduces parameter updates, leverages pre-trained knowledge more effectively, and supports robust generalization across diverse datasets.
}

Our primary contributions include:
\begin{itemize}
    \item Introducing a parameter-efficient fine-tuning strategy that retains and leverages both encoder and decoder from the pre-training stage, effectively mitigating catastrophic forgetting.
    \item Developing novel fine-tuning modules (CEP, MCP, and PA) to enhance the adaptability of motion forecasting models while minimizing retrainable parameters.
    \item Conducting extensive experiments, including comprehensive ablation studies within single-dataset and cross-dataset settings, demonstrating the efficacy, scalability, and practical applicability of our Forecast-PEFT approach.
\end{itemize}

Through these contributions, Forecast-PEFT emerges as a robust and scalable solution for motion prediction in autonomous driving, offering substantial computational efficiency and improved generalization capabilities compared to conventional fine-tuning methods \cite{wilson2023argoverse}.

\section{Related Work}
\subsubsection{
Self-Supervised Pretraining in Motion Prediction} 
\revise{Trajectory prediction has evolved from early social interaction models 
\cite{Alahi_2016_CVPR, Gupta_2018_CVPR} to graph-based and transformer-based architectures 
\cite{9084255, 9576054, 8813829, saadatnejad2024socialtransmotion}.} 
\revise{Representative methods such as MTR++~\cite{shi2024mtr++}, QCNet~\cite{zhou2023query}, and Wayformer~\cite{nayakanti2023wayformer} achieve strong performance through query-centric or transformer-based architectures with efficient attention mechanisms, while recent work further advances prediction accuracy through historical attention~\cite{tang2024hpnet} and causal reasoning~\cite{pourkeshavarz2024cadet}.}
\revise{While recent frameworks such as UniTraj~\cite{feng2024unitraj} provide unified architectures for scalable trajectory prediction across datasets, the efficient adaptation of pre-trained models remains under-explored, which is the focus of our work.}  
In parallel, studies on self-supervised pretraining methods for motion prediction have achieved remarkable success. These approaches adopt pretext tasks to pretrain foundation models to learn latent semantic information of both map and motion of road users, and then fine-tune the model on motion prediction task. Existing pre-training methods can be divided into contrastive representation learning \cite{liu2021social, xu2022pretram, bhattacharyya2023ssl, pourkeshavarz2023learn, chib2023enhancing, azevedo2022exploiting} and generative masked representation learning \cite{gao2020vectornet, chen2023traj, li2023pre, cheng2023forecast, lan2023sept, Messaoud_2025_CVPR}. The contrastive learning approaches align and distinguish representations of views with the same high-level semantics or not. Social-NCE \cite{liu2021social} employs contrastive loss and negative sample augmentation to extract information about social interactions from motion trajectories. PreTraM \cite{xu2022pretram} and MENTOR \cite{pourkeshavarz2023learn} explore the use of contrast learning of intra-domain (map-map) and cross-domain (trajectory-map) features. 
Generative masked pretraining methods mainly rely on the masked autoencoders \cite{he2022masked} to learn latent representation by reconstructing the masked input. Traj-MAE \cite{chen2023traj} and SEPT \cite{lan2023sept} explore to pretrain the map and motion encoder separately with specially designed mask-reconstruction strategy. 
Forecast-MAE \cite{cheng2023forecast} takes agents' trajectory and poly-lines of lane segments as single tokens, and then randomly masks out the entire agents’ history or future trajectory, as well as lane segments, at the token level, before feeding them into the Transformer \cite{vaswani2017attention} backbone. \revise{T4P~\cite{park2024t4p} extends this approach to test-time adaptation, while SmartPretrain~\cite{zhou2025smartpretrain} proposes model-agnostic and dataset-agnostic pretraining for improved cross-dataset generalization.} 
Despite the success of pretraining in motion prediction, two issues have not been fully explored. Firstly, all methods in the pretraining stage utilize a decoder that is subsequently discarded. During the fine-tuning stage for motion prediction, a new decoder must be designed specifically for prediction purposes. Secondly, the adaptation for motion prediction still demands resource-intensive full fine-tuning, and effective Parameter-Efficient Fine-Tuning (PEFT) methods have yet to be explored. Thus, we explore both the re-use of pre-trained decoder and PEFT techniques in the motion prediction domain. 

\subsubsection{Parameter-efficient Fine-tuning} Parameter-efficient fine-tuning (PEFT) aims to fine-tune only a small set of parameters, which might be a set of newly added parameters \cite{jia2022visual, lialin2023scaling} or a small subset of the pre-trained model's parameters. 
\revise{Recent work demonstrates that such parameter-efficient adaptations can reach competitive or even superior performance compared to full fine-tuning, while substantially reducing training and deployment costs~\cite{yin20255}. Comprehensive studies in visual 
recognition~\cite{Mai_2025_CVPR} further analyze the effectiveness of different PEFT strategies across diverse tasks.} 
Existing PEFT methods can be primarily divided into additive methods and selective methods, along with hybrid versions of these methods. The additive methods, which include approaches like Prompt Tuning \cite{jia2022visual}, Adapter Tuning \cite{houlsby2019parameter, pfeiffer2020adapterhub, pfeiffer2020adapterfusion, chen2022adaptformer, he2021towards}, LoRA \cite{hu2021lora}, Prefix Tuning\cite{li2021prefix}, and Side Tuning \cite{zhang2020side}, involve adding new trainable parameters or modules to pre-trained models for efficient fine-tuning while preserving original parameters. The selective methods, such as LayerNorm-Tuning \cite{qi2022parameter}, Bias Tuning \cite{zaken2021bitfit, cai2020tinytl}, and strategies like Partial-k \cite{he2022masked, yosinski2014transferable}, focus on tuning a subset of existing model parameters, targeting those critical for specific tasks to reduce computational load. Additionally, there is an increasing trend towards integrating these methods, aiming to combine the flexibility of additive methods with the targeted efficiency of selective methods for optimal model performance and resource efficiency. Recent works explore hybrid PEFT strategies, such as instance-aware dynamic prompt tuning (IDPT)~\cite{zha2023instance} and the combination of dynamic adapters with prompt tuning (DAPT)~\cite{zhou2024dynamic}, demonstrating enhanced parameter efficiency in point cloud models. However, these approaches are designed primarily for point cloud analysis rather than general motion forecasting. The recent work MoSA~\cite{kothari2023motion} introduces a modular low-rank adapter for motion style transfer, but only for adapting motion forecasting models pre-trained on one domain with sufficient data to new domains such as unseen agent types and scene contexts. 
In this paper, we propose a PEFT approach for the general motion prediction domain, which includes contextual embedding prompts, modality control prompts, and parallel adapters. 

\begin{figure*}[t]
\begin{center}
\includegraphics[width=1.0\linewidth]{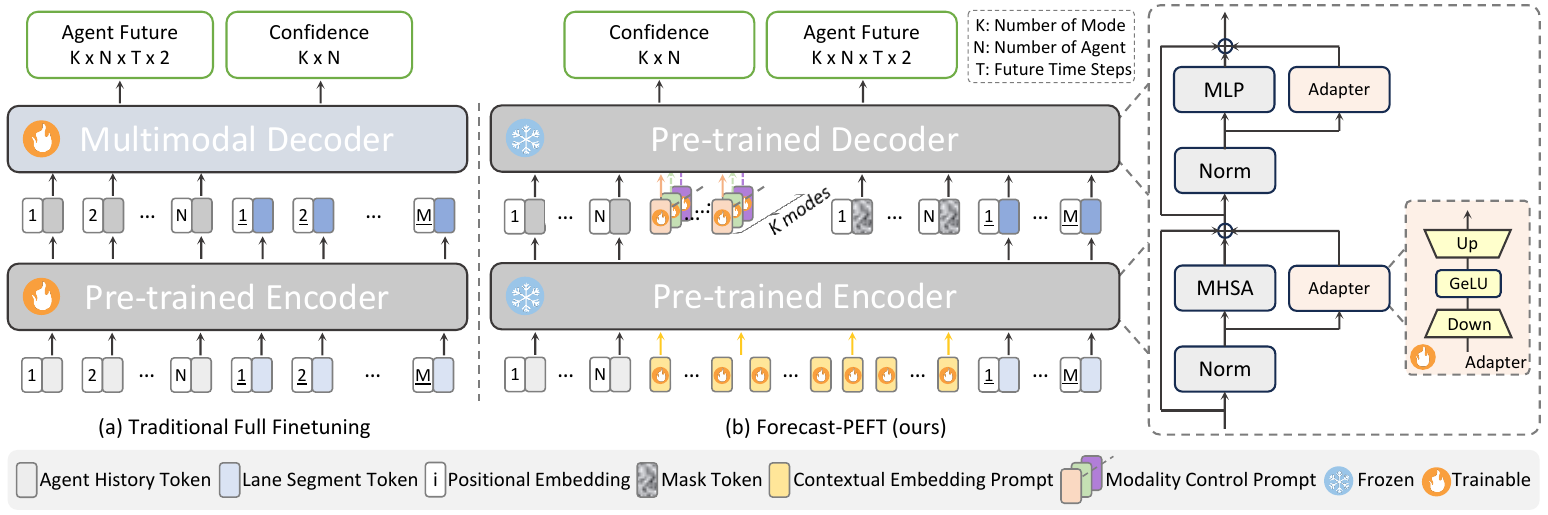}
\end{center}
   \caption{   \textbf{Forecast-PEFT \textit{vs.} traditional full finetuning}: (a) Traditional methods, exemplified by Forecast-MAE \cite{cheng2023forecast}, use a pre-trained encoder and attach a randomly initialized new decoder. (b) Forecast-PEFT retains the pre-trained decoder, initially for masked motion/map reconstruction, adapting it for future motion forecasting by integrating Modality Control Prompts into the decoder, Contextual Embedding Prompts into the encoder, and incorporating Parallel Adapters.
   }
\label{fig:Forecast-PEFT}
\end{figure*}

\section{Method}

In this section, we first introduce the standard fine-tuning pipeline for a pre-trained motion prediction model in Section \ref{sec:section3-1}, followed by the background concept of Visual Prompt Tuning in Section \ref{sec:section3-vpt}. Then, in Section \ref{sec:section3-2} and \ref{sec:section3-3}, we describe our Forecast-PEFT and Forecast-FT frameworks, respectively.

\subsection{Standard Fine-Tuning Pipeline in Forecast-MAE}
\label{sec:section3-1}
\subsubsection{Overview of Forecast-MAE}
Forecast-MAE \cite{cheng2023forecast} uses a self-supervised MAE-based \cite{he2022masked} method for motion forecasting, uniquely combining agent trajectory and random lane segment masking during pre-training to understand the intricate relationship between road geometry and agent dynamics.

First, agent trajectories and lane segments are transformed into vectorized tokens. These tokens undergo processing using a Feature Pyramid Network (FPN)~\cite{lin2017feature} for $N$ agents (\(T_H = FPN(A_H), T_F = FPN(A_F)\), where $T_{H,F} \in \mathbb{R}^{N\times C}$) and Mini-PointNet~\cite{qi2017pointnet} for $M$ lane segments (\(T_L = MiniPointNet(L)\), where $T_{L} \in \mathbb{R}^{M\times C}$). $T_H, T_F, T_L$ are history, future, lane tokens respectively, and $C$ is the embedding dimension. Additionally, semantic and positional embeddings (PE) are incorporated into these tokens. Notably, Forecast-MAE masks either the future or history trajectories of agents in a complementary manner. This means that if the history trajectory of an agent is masked, its future trajectory remains unmasked, and vice versa.

Then, the architecture employs a standard Transformer-based \cite{vaswani2017attention} autoencoder. The encoder and decoder processes are formulated as:
\begin{align}
    T_E &= Encoder\big(concat\left(T_H, T_F, T_L\right) + PE\big), \\ 
    M' &= Decoder\big(concat\left(T_E,M\right) + PE\big),
\end{align}
where $M=(M_H,M_F,M_L)$ are the mask tokens concatenated to the encoded latent tokens of the encoder, and $M'=(M_H^{'}, M_F^{'},M_L^{'})$ are the decoded mask tokens.   

The decoder's role is to reconstruct masked elements, predicting normalized coordinates for masked trajectories and lane segments, with reconstruction targets processed through: 
\begin{align}
    P_i &= PredictionHead_i(M_i'), i\in\{H, F, L\}. 
\end{align}

\subsubsection{Pre-training Loss} The model utilizes the L1 loss terms $\mathcal{L}_H$ and $\mathcal{L}_F$ for trajectory reconstruction and the mean squared error (MSE) loss $\mathcal{L}_L$ for lane segment reconstruction. $\lambda_H$, $\lambda_F$, and $\lambda_L$ are the weights of the loss terms. The final loss, denoted as $\mathcal{L}_{RE}$ and defined in the following equation, is also employed to represent the Reconstruction Error (RE) in Figure \ref{fig:start}, thereby quantifying the level of catastrophic forgetting:
\begin{align}
\begin{split}
    \mathcal{L}_{RE} = \lambda_H \mathcal{L}_H + \lambda_F \mathcal{L}_F + \lambda_L \mathcal{L}_L.
\end{split}
\end{align}

\subsubsection{Model Adapting and Fine-tuning} In fine-tuning for motion forecasting, the MAE-specific Decoder is replaced with a randomly initialized MLP-based multi-modal decoder that generates various future trajectory forecasts, with the entire architecture undergoing full fine-tuning:
\begin{align}
    T_E' &= Encoder\big(concat\left(T_H, T_L\right) + PE\big), \\ 
    P_{F} &= MultiModalDecoder\left(T_E'\right), P_{F} \in \mathbb{R}^{N\times K \times T \times 2}, 
    \label{eq:MultiModalDecoder}
\end{align}
where $T$ is the number of future frames, $K$ is the number of output modes.

\revise{\subsection{Visual Prompt Tuning (VPT)}
\label{sec:section3-vpt}
VPT is a parameter-efficient fine-tuning method that introduces learnable prompt tokens into the input space, enabling adaptation without modifying model weights \cite{jia2022visual}. It has been widely applied in vision-language and classification tasks. VPT can be categorized into two variants: VPT-Deep, which integrates prompts across all Transformer layers for deeper feature refinement, and VPT-Shallow, which inserts prompts only at the initial layers for lightweight tuning \cite{jia2022visual}. The design of our prompts is inspired by VPT, adapting its principles to the motion forecasting task.
}

\subsection{Forecast-PEFT}
\label{sec:section3-2}

The pre-trained Forecast-MAE model, depicted in Figure \ref{fig:Forecast-PEFT}, is enhanced by Forecast-PEFT through the integration of specialized prompts and adapters as additive modules, fine-tuned to enhance trajectory prediction tasks.

\revise{The design of these modules follows a unified principle: the largely frozen pre-trained model is adapted through lightweight trainable components operating on the encoder input, decoder generation, and intermediate representations during the transition from masked pre-training to downstream forecasting. Specifically, CEP addresses the input-space mismatch between masked trajectory--map inputs during pre-training and history-conditioned inputs during forecasting; MCP addresses the output-task mismatch by conditioning the pre-trained decoder to generate multiple future modes instead of reconstructing masked tokens; and PA refines intermediate representations through bottleneck residual transformations while most backbone parameters remain frozen. Their complementary effects are evaluated in the ablation study in Section~\ref{Ablation_Study}.}

\subsubsection{Contextual Embedding Prompt (CEP)}
The Forecast-PEFT framework incorporates CEP vectors denoted as $P_{CE}$ to augment the pre-trained Encoder's contextual comprehension of input data. 
\revise{Similar to VPT-Deep \cite{jia2022visual}, CEPs are introduced at the input of each Transformer layer.} Applying deep tuning in the encoder allows for more nuanced and complex feature extraction from the input data. It enhances the encoder's ability to understand and encode subtle patterns and relationships within the data, setting a strong foundation for the decoding process. The encoder process is formulated as:
\begin{align}
    T_E' &= Encoder\big(concat\left(T_H, \boldsymbol{P_{CE}}, T_L\right) + PE\big),
\end{align}
where $P_{CE} \in \mathbb{R}^{N_P \times C \times L_E}$, and $N_P, C, L_E$ denote \revise{the prompt length}, embedding dimension, and number of the Transformer encoder layers, respectively. \revise{This means that layer-specific CEPs are inserted into the token sequence between history and map tokens at each Transformer encoder layer, as illustrated in Figure~\ref{fig:Forecast-PEFT}. These tokens then participate in the multi-head self-attention. Positional embeddings are added only at the first layer, following standard practice for Transformers.}

\subsubsection{Modality-Control Prompt (MCP)} The pre-trained decoder's capability is expanded by MCPs to handle multi-modal output generation.
Similar to VPT Shallow \cite{jia2022visual}, MCP vectors, denoted as $P_{MC}$, are inserted only at the level of the first Transformer layer of the decoder. We adopt VPT Shallow here since each MCP set is designed to guide the decoder in generating a distinct mode of future trajectories. The decoder finetuning process is formulated as:
\begin{align}
    M_F' &= Decoder\big(concat\left(T_E', \boldsymbol{P_{MC}}, M_F\right)+PE\big), \\
    P_F &= PredictionHead_F\left(M_F'\right), P_{F} \in \mathbb{R}^{N\times K \times T \times 2}, 
\end{align}
where $P_{MC} \in \mathbb{R}^{N_P \times C \times K}$, and $N_P, C, K$ denote \revise{the prompt length}, embedding dimension, and number of \revise{prediction} modes, respectively. 
In this way, the pre-trained decoder is reused for generating multi-modal future trajectories. \revise{As illustrated in Figure~\ref{fig:Forecast-PEFT}, MCP vectors are concatenated with the encoder output and masked future tokens at the decoder's first layer. They participate in the multi-head self-attention as well, with each MCP guiding the generation of one prediction mode. 
The K modes can be processed sequentially or in parallel within a single forward pass by expanding the decoder input along the mode dimension (Section~IV-G).
}

\subsubsection{Parallel Adapter (PA)}

Complementing CEPs and MCPs, we introduce parallel adapters (PAs) to enable fine-grained local feature adaptation. \revise{Our adapters consist of a fully connected layer for downscaling (Down), followed by a GeLU activation function, and then a fully connected layer for upscaling (Up):}
\begin{align}
    \boldsymbol{Adapter}(f_{input}) &= Up\big(GeLU(Down(f_{input}))\big),
\end{align}
where $f_{input}$ is the \revise{layer-normalized} input feature.
\revise{Here, $Down$ corresponds to $W_{\text{down}} \in \mathbb{R}^{C \times r}$ and $Up$ to $W_{\text{up}} \in \mathbb{R}^{r \times C}$, where $C$ is the embedding dimension and the bottleneck size $r$ defines the rank, balancing adaptation capacity and parameter efficiency.} \revise{As illustrated in Figure~\ref{fig:Forecast-PEFT}, parallel adapters are inserted alongside both MHSA and FFN modules within each Transformer block. Each adapter processes the same layer-normalized input as its main module through this bottleneck structure, with outputs combined as $x' = x + \text{Module}(\text{norm}(x)) + \text{Adapter}(\text{norm}(x))$.}

\subsubsection{Selective Unfreezing} Above is our design and utilization of additive PEFT modules. For selective PEFT modules, we also unfreeze the Bias, LayerNorm and PredictionHead of the pre-trained model, which have a limited number of tunable parameters -- 18.9K, 3.7K, and 31.7K, respectively. 

\subsubsection{Variants} \revise{We evaluate two variants. The first variant, which only tunes the additive modules (prompts and adapters), is denoted by Forecast-PEFT (A). The second variant, which is the default variant and denoted by Forecast-PEFT, includes both additive modules and selective unfreezing.} 

\subsubsection{Loss for Fine-tuning} The target agent's loss is calculated using Huber loss for trajectory regression and cross-entropy loss for confidence classification. A winner-takes-all (WTA) approach determines the regression loss by comparing the best prediction, chosen based on the lowest average displacement error across all timesteps, with the ground truth.
\subsubsection{Cross-Dataset Finetuning}
Our Forecast-PEFT explores another way to make the best use of the general knowledge of large datasets, which is using pre-training and parameter-efficient fine-tuning, instead of transfer learning. Our additive modules and selective modules can be seen together as a small plug-in module, and we only let this module learn domain-specific knowledge.

\begin{table}[tb!]
\caption{\textbf{Quantitative Results on the AV2 test set.} The percentage shows the improvement of Forecast-FT compared to Forecast-MAE. 
}
\label{tab:leaderboard}
\centering
    \resizebox{1.0\linewidth}{!}{
  \fontsize{22pt}{18pt} \selectfont
  \begin{tabular}{lllll}
    \toprule
    \textbf{Method} & \textbf{minADE ($\downarrow$)} & \textbf{minFDE ($\downarrow$)} & \textbf{MR ($\downarrow$)} & \textbf{b-minFDE} ($\downarrow$) \\ 
    \midrule
    SSL-Lanes~\cite{bhattacharyya2023ssl} & \hspace{0.4cm}0.83 & \hspace{0.3cm}1.67 & \hspace{0.1cm}0.25 & \hspace{0.5cm}2.35 \\
    GoReLa~\cite{cui2022gorela} & \hspace{0.4cm}0.76 & \hspace{0.3cm}1.48 & \hspace{0.1cm}0.22 & \hspace{0.5cm}2.01 \\
    GANet~\cite{wang2022ganet} & \hspace{0.4cm}0.73 & \hspace{0.3cm}{1.35} & \hspace{0.1cm}0.17 & \hspace{0.5cm}1.97 \\
    Gnet~\cite{gao2023dynamic} & \hspace{0.4cm}0.69 & \hspace{0.3cm}1.34 & \hspace{0.1cm}0.18 & \hspace{0.5cm}1.90 \\
    Macformer~\cite{feng2023macformer} & \hspace{0.4cm}0.70 & \hspace{0.3cm}1.38 & \hspace{0.1cm}0.19 & \hspace{0.5cm}1.90 \\
    ProphNet~\cite{wang2023prophnet} & \hspace{0.4cm}0.66 & \hspace{0.3cm}1.32 & \hspace{0.1cm}0.18 & \hspace{0.5cm}1.88 \\
    QCNet~\cite{zhou2023query} & \hspace{0.4cm}0.62 & \hspace{0.3cm}1.19 & \hspace{0.1cm}0.14 & \hspace{0.5cm}1.78 \\
    SEPT~\cite{lan2023sept} & \hspace{0.4cm}0.61 & \hspace{0.3cm}1.15 & \hspace{0.1cm}0.14 & \hspace{0.5cm}1.74 \\
    MTR$+$$+$~\cite{shi2024mtr++}& \hspace{0.4cm}0.71 & \hspace{0.3cm}1.37 & \hspace{0.1cm}0.14 & \hspace{0.5cm}1.88 \\
    Forecast-MAE w/esb\footnotemark & \hspace{0.4cm}0.690 & \hspace{0.3cm}1.338 & \hspace{0.1cm}0.173 & \hspace{0.5cm}1.911 \\
    \midrule
    Forecast-MAE~\cite{cheng2023forecast}  & \hspace{0.4cm}0.716 & \hspace{0.3cm}1.412 & \hspace{0.1cm}0.174 & \hspace{0.5cm}2.050 \\
    \baseline{Forecast-PEFT (ours)} & \hspace{0.4cm}\baseline{0.707} & \hspace{0.3cm}\baseline{1.367} & \hspace{0.1cm}\baseline{0.172} & \hspace{0.5cm}\baseline{2.016} \\ 
    \baseline{Forecast-FT (ours)} & \hspace{0.4cm}\baseline{0.687 \red{$\downarrow$4.0\%}} & \hspace{0.3cm}\baseline{1.309 \red{$\downarrow$7.3\%}} & \hspace{0.1cm}\baseline{0.157 \red{$\downarrow$9.6\%}} & \hspace{0.5cm}\baseline{1.960 \red{$\downarrow$4.4\%}} \\
    \bottomrule
    \end{tabular}}
\end{table}

\footnotesize
\vspace{-0.8em}
\footnotetext{w/esb refers to ``with ensemble".}
\vspace{-1.0em}
\normalsize

\subsection{Forecast-FT}
\label{sec:section3-3}
The Forecast-FT framework is closely aligned with the Forecast-PEFT approach, differing primarily in the extent of fine-tuning. While Forecast-PEFT focuses on a parameter-efficient fine-tuning strategy by integrating specialized prompts and adapters and selectively fine-tuning parts of the model, Forecast-FT extends this concept by engaging in a more comprehensive fine-tuning process. Specifically, Forecast-FT removes the Parallel Adapters integrated into the Forecast-PEFT framework and proceeds to fully fine-tune all parameters of the pre-trained model. This approach aims to harness the full potential of the pre-trained model, adjusting all its parameters to enhance its predictive accuracy.

\section{Experiments}
To evaluate the effectiveness of our proposed Forecast-PEFT framework, we conducted a series of comprehensive experiments across multiple motion forecasting datasets. Section \ref{Experiment_Setup} provides a detailed overview of our experimental setup. A thorough analysis of the performance of Forecast-PEFT on individual datasets is presented in Sections \ref{Quantitative_Analysis} and \ref{Qualitative_Analysis}. We examine its ability to preserve pre-trained knowledge in Section \ref{Preservation_Knowledge} and its generalization capabilities across different datasets in Section \ref{Cross_Dataset}. Additionally, the efficiency of Forecast-PEFT in terms of training time, data usage, and storage requirements is investigated in Section \ref{Efficiency_Analysis}. Finally, we present an ablation study to assess the contributions of various components within the Forecast-PEFT framework in Section \ref{Ablation_Study}. Our experiments demonstrate the advantages of Forecast-PEFT in achieving high prediction accuracy with significantly fewer trainable parameters compared to traditional fine-tuning methods.

\begin{table}[t]
\caption{\textbf{Quantitative Results on the AV1 test set.} 
}
\vspace{1.3em}
\label{tab:argo_results}
\centering
    \resizebox{1.0\linewidth}{!}{
    \fontsize{22pt}{18pt} \selectfont
  \begin{tabular}{lllll}
    \toprule
    \textbf{Method} & \textbf{minADE ($\downarrow$)} & \textbf{minFDE ($\downarrow$)} & \textbf{MR ($\downarrow$)} & \textbf{b-minFDE} ($\downarrow$) \\ 
    \midrule
    THOMAS ~\cite{gillesthomas} & \hspace{0.4cm}0.94 & \hspace{0.3cm}1.44 & \hspace{0.1cm}0.104 & \hspace{0.5cm}1.97 \\
    SSL-Lanes~\cite{bhattacharyya2023ssl} & \hspace{0.4cm}0.84 & \hspace{0.3cm}1.25 & \hspace{0.1cm}0.13 & \hspace{0.5cm}1.94 \\
    AutoBots~\cite{girgis2021latent} & \hspace{0.4cm}0.89 & \hspace{0.3cm}1.41 & \hspace{0.1cm}0.16 & \hspace{0.5cm}2.06 \\ 
    GANet ~\cite{wang2022ganet} & \hspace{0.4cm}0.81 & \hspace{0.3cm}1.16 & \hspace{0.1cm}0.12 & \hspace{0.5cm}1.79 \\
    Wayformer~\cite{nayakanti2023wayformer} & \hspace{0.4cm}0.77 & \hspace{0.3cm}1.16 & \hspace{0.1cm}0.12 & \hspace{0.5cm}1.74 \\
    Gnet~\cite{gao2023dynamic}& \hspace{0.4cm}0.79 & \hspace{0.3cm}1.16 & \hspace{0.1cm}0.12 & \hspace{0.5cm}1.75 \\
    Macformer~\cite{feng2023macformer} & \hspace{0.4cm}0.81 & \hspace{0.3cm}1.21 & \hspace{0.1cm}0.13 & \hspace{0.5cm}1.77 \\
    ProphNet~\cite{wang2023prophnet} & \hspace{0.4cm}0.76 & \hspace{0.3cm}1.13 & \hspace{0.1cm}0.11 & \hspace{0.5cm}1.69 \\
    QCNet~\cite{zhou2023query} & \hspace{0.4cm}0.73 & \hspace{0.3cm}1.07 & \hspace{0.1cm}0.106 & \hspace{0.5cm}1.69 \\
    SEPT~\cite{lan2023sept} & \hspace{0.4cm}0.73 & \hspace{0.3cm}1.06 & \hspace{0.1cm}0.103 & \hspace{0.5cm}1.68 \\
    \midrule
    Forecast-MAE~\cite{cheng2023forecast} & \hspace{0.4cm}0.870 & \hspace{0.3cm}1.348 & \hspace{0.1cm}0.151 & \hspace{0.5cm}2.041 \\ 
    \baseline{Forecast-PEFT (ours)} & \hspace{0.4cm}\baseline{0.856} & \hspace{0.3cm}\baseline{1.294} & \hspace{0.1cm}\baseline{0.146} & \hspace{0.5cm}\baseline{1.976} \\ 
    \baseline{Forecast-FT (ours)} & \hspace{0.4cm}\baseline{0.842 \red{$\downarrow$3.2\%}} & \hspace{0.3cm}\baseline{1.270 \red{$\downarrow$5.8\%}} & \hspace{0.1cm}\baseline{0.138 \red{$\downarrow$8.6\%}} & \hspace{0.5cm}\baseline{1.957 \red{$\downarrow$4.1\%}} \\ 
    \bottomrule
    \end{tabular}
     }
\end{table}

\subsection{Experiment Setup}
\label{Experiment_Setup}
\subsubsection{Datasets} For our experiments, we utilize the Argoverse 2 \cite{wilson2023argoverse} (AV2) and Argoverse 1 (AV1) \cite{chang2019argoverse} motion forecasting datasets, while the nuScenes dataset \cite{caesar2020nuscenes} is used only for cross-dataset evaluations.\\
The \textbf{AV2 dataset} \cite{wilson2023argoverse} provides 250,000 real-world scenarios, divided into 199,908 training samples, 24,988 validation samples, and 24,984 test samples. Each scenario features high-definition maps and 11-second trajectories sampled at 10 Hz, with 5 seconds dedicated to observed history and 6 seconds for future forecasting.\\
The \textbf{AV1 dataset} \cite{chang2019argoverse} contains 324,000 scenarios, which are split into 205,942 training samples, 39,472 validation samples, and 78,143 test samples. This dataset includes high-definition maps and 5-second trajectories also sampled at 10 Hz, providing 2 seconds of history and 3 seconds of future data.\\
The \textbf{nuScenes dataset} \cite{caesar2020nuscenes} offers high-definition semantic maps and 8-second trajectories, with 2 seconds for observed history and 6 seconds for future forecasting. Unlike the Argoverse datasets, nuScenes has a lower sampling rate of 2 Hz. It consists of 1000 driving scenes, split into 700 training scenes, 150 validation scenes, and 150 test scenes.\\
Figure \ref{fig:cross-dataset} illustrates the differences in sampling rates and trajectory lengths across these datasets, emphasizing the distinct characteristics and challenges each dataset presents for cross-dataset evaluation.

\subsubsection{Metrics} The metrics that we use are from the official AV2 \cite{wilson2023argoverse} and AV1 \cite{chang2019argoverse} benchmarks. We use minimum Average Displacement Error (minADE), minimum Final Displacement Error (minFDE), missing rate (MR), and Brier minimum Final Displacement Error (b-minFDE) for six prediction modes unless specified otherwise. 

\subsubsection{Implementation details}
Our base model is Forecast-MAE \cite{cheng2023forecast}, featuring a pre-trained encoder and decoder with 4 transformer layers each. We adopt a balanced setting for reconstruction losses in pre-training with $\lambda_H = \lambda_F = 1.0$ and $\lambda_L = 0.35$. The model has feature dimension $C = 128$ and we use six modes ($K=6$), with CEPs and MCPs lengths set at $N_P=50$. CEPs match the encoder's 4-layer depth, while MCPs, inserted only in the first decoder layer, feature 6 parallel prompts, corresponding to the number of modes. Adapters (PAs) have a hidden size of $r=64$. For both Forecast-PEFT and Forecast-FT, we train for 60 epochs using AdamW  \cite{loshchilov2017decoupled}, a batch size of 128 across 4 NVIDIA A100 GPUs, weight decay of 1e-4, and an initial learning rate of 1e-3 with cosine annealing decay. 
Prompts are initialized using the Xavier uniform scheme \cite{glorot2010understanding}, while all adapter weights and biases are set to zero, maintaining the backbone model's state as identical to the pre-trained version.

\subsection{Quantitative Analysis}
\label{Quantitative_Analysis}
\subsubsection{Performance of Forecast-PEFT}
Tables \ref{tab:leaderboard} and \ref{tab:argo_results} show that our Forecast-PEFT framework surpasses the full fine-tuning baseline Forecast-MAE (FT) over all metrics on both AV2 and AV1 datasets, with only 20\% of the trainable parameter. Our Forecast-PEFT (A), as shown in Table \ref{tab:compare_peft}, achieves comparable results as the baseline with only 17\% of the trainable parameters. These results consistently indicate the parameter efficiency of our framework and the effectiveness of our additive and selective modules. 

\subsubsection{Performance of Forecast-FT} Table \ref{tab:leaderboard} and \ref{tab:argo_results} on both leaderboards of AV2 and AV1 show that our Forecast-FT surpasses the original full fine-tuning baseline method \cite{cheng2023forecast} for at least 4\% over all metrics. Specifically, Forecast-FT achieves an improvement of MR by 9.6\% and 8.6\% on AV2 and AV1, respectively. These results indicate the advantage and effectiveness of our design by using both the pre-trained encoder and decoder. 

\begin{figure*}[ht!]
\begin{center}
\includegraphics[width=1.005\linewidth]{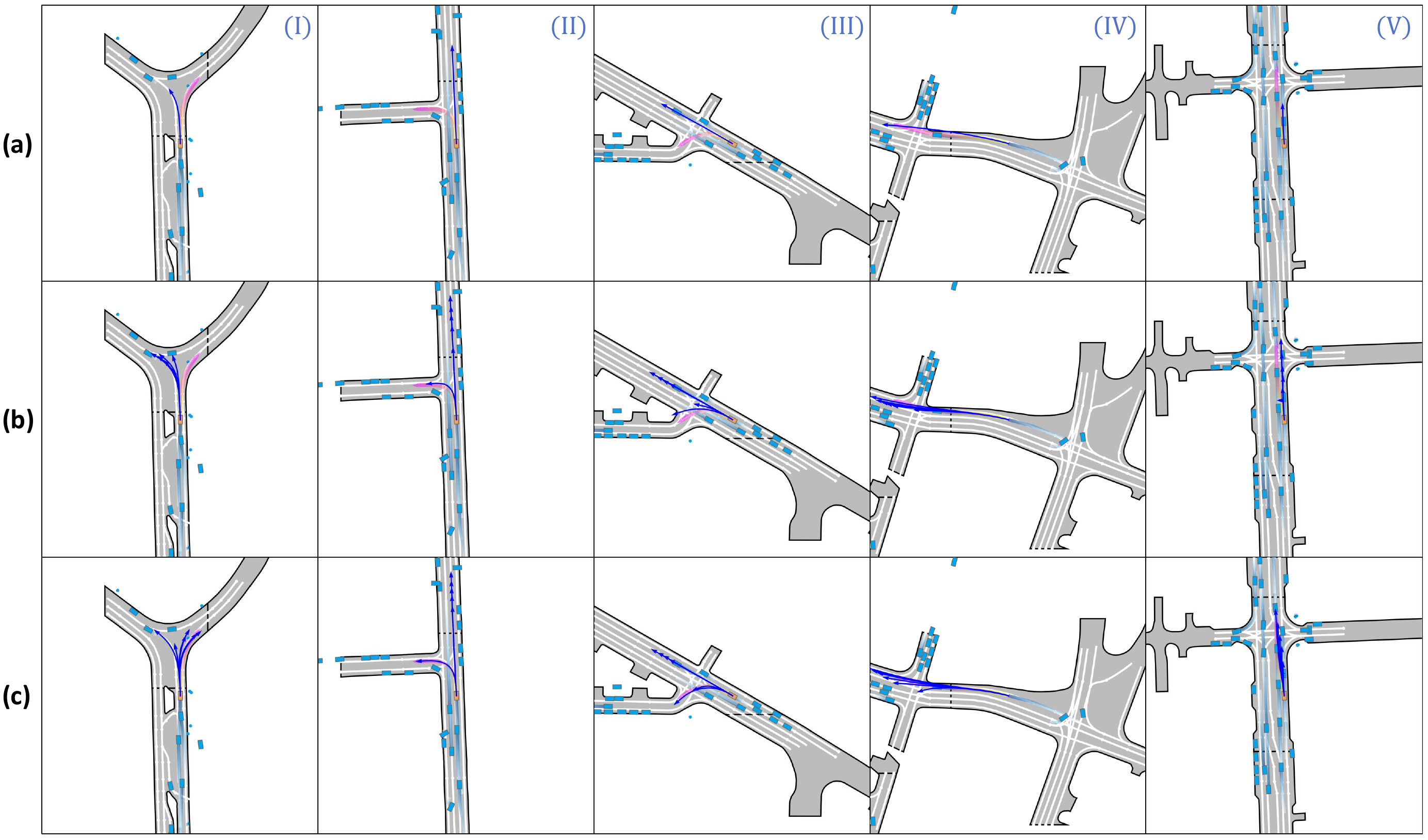}
\end{center}
   \caption{
   \textbf{Qualitative results} of (a) pre-trained encoder and decoder of Forecast-MAE, (b) our proposed Forecast-PEFT, and (c) Forecast-FT. The history trajectories are in light blue, the forecasted future trajectories are in dark blue, and the ground truth is in gradient pink. The box in orange indicates the target agent, and the box in blue denotes neighbor agents. 
   }
\label{fig:qualitative}
\end{figure*}

\subsubsection{Comparison with other PEFT methods} 
\label{Comparison_PEFT}
 \revise{As no specific PEFT method exists for motion forecasting, we benchmark our approach against several techniques from other domains. Specifically, we compare with VPT \cite{jia2022visual} (both shallow and deep variants) from vision, LoRA \cite{hu2021lora} from language, IDPT \cite{zha2023instance} from 3D point clouds, as well as with Prefix Tuning \cite{li2021prefix}, MoSA \cite{kothari2023motion}, and DAPT \cite{zhou2024dynamic} that adjust the original multi-modal decoder (MD). All these methods retain the original MD to integrate their prompts or adapters. As shown in Table \ref{tab:compare_peft}, our Forecast-PEFT framework significantly outperforms these techniques. In particular, Forecast-PEFT (A) achieves a minADE of 0.704, minFDE of 1.373, and MR of 0.181 using only 329K parameters, while the full Forecast-PEFT further reduces the errors to 0.696 (minADE) and 1.351 (minFDE) with an MR of 0.173 using 383K parameters—delivering competitive performance with only a fraction of the 1.9M parameters used in full fine-tuning.}

\subsubsection{Choosing Forecast-PEFT for Efficiency and Flexibility} 
Although our fully fine-tuned variant Forecast-FT demonstrates the effectiveness of our approach without any design changes, we choose Forecast-PEFT for its efficiency and flexibility. Forecast-PEFT tunes only 15\% (0.38M) of the trainable parameters required by Forecast-FT (2.5M) while maintaining good performance. 

\begin{table}[b!]
\caption{\textbf{Comparison with different PEFT methods on the AV2 validation set.} The Multi-modal Decoder is the one we described in Equation \ref{eq:MultiModalDecoder}, which is the original decoder of our baseline \cite{cheng2023forecast} finetuned to get multi-modal output.
}
\label{tab:compare_peft}
\tablestyle{8pt}{1.1}
\resizebox{1.0\linewidth}{!}{
\fontsize{12pt}{12pt} \selectfont
\begin{tabular}{ccccc}
\toprule
Method & Params & minADE & minFDE & MR \\ \midrule
Forecast-MAE~\cite{cheng2023forecast} & 1.9M & 0.711 & 1.406 & 0.178 \\
\midrule
Multi-modal Decoder & 246K & 0.905 & 1.982 & 0.327 \\ 
VPT Shallow \cite{jia2022visual} & 252K & 0.852 & 1.814 & 0.286 \\
VPT Deep \cite{jia2022visual} & 272K & 0.843 & 1.793 & 0.281 \\
LoRA (r=16) \cite{hu2021lora} & 279K & 0.828 & 1.762 & 0.26 \\ 
LoRA (r=64) \cite{hu2021lora} & 377K & 0.786 & 1.638 & 0.235 \\ 
IDPT \cite{zha2023instance} & 395K & 0.809 & 1.669 & 0.255 \\ 
\revise{Prefix Tuning \cite{li2021prefix} } & \revise{318K} & \revise{0.835} & \revise{1.753} & \revise{0.269} \\ 
\revise{MoSA \cite{kothari2023motion} } & \revise{312K} & \revise{0.773} & \revise{1.567} & \revise{0.226} \\ 
\revise{DAPT \cite{zhou2024dynamic} }  & \revise{326K} & \revise{0.768} & \revise{1.554} & \revise{0.223} \\ \midrule
\baseline{Forecast-PEFT (A)} & \baseline{329K} & \baseline{0.704} & \baseline{1.373} & \baseline{0.181} \\
\baseline{Forecast-PEFT} & \baseline{383K} & \baseline{0.696} & \baseline{1.351} & \baseline{0.173} \\ \bottomrule
\end{tabular}
}
\end{table}

\subsection{Qualitative Analysis}
\label{Qualitative_Analysis}
In Figure \ref{fig:qualitative}, we display qualitative results from AV2 using pre-trained Forecast-MAE, and our fine-tuned versions, Forecast-PEFT and Forecast-FT. The pre-trained Forecast-MAE's encoder and decoder demonstrate reasonable single-mode predictions, supporting our assertion that the pre-trained decoder is suitable for direct future trajectory prediction. Forecast-PEFT, leveraging the pre-trained components while minimally adjusting parameters, achieves diverse, multi-modal predictions. Forecast-FT, with all pre-trained weights unfrozen, shows enhanced trajectory prediction accuracy, as confirmed by the visualization results.

\begin{figure*}[ht!]
\begin{center}
\includegraphics[width=1.0\linewidth]{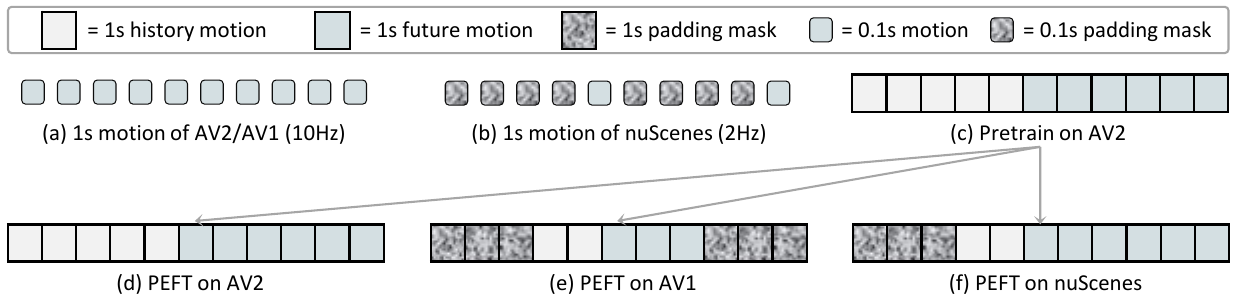}
\end{center}
   \caption{\textbf{Standardization techniques used for cross-dataset fine-tuning.} (a) AV2 and AV1 datasets sampled at 10Hz provide detailed historical and future trajectories, while (b) nuScenes dataset sampled at 2Hz offers sparse trajectories. (c) Pre-training is performed on the AV2 dataset with masked trajectory inputs. \revise{(d-e) For fine-tuning on datasets with shorter trajectory lengths (AV1) or lower sampling rates (nuScenes), missing data points are represented as masked entries (zero padding with validity masks) to standardize inputs.} (f) Forecast-PEFT effectively adapts the model across these diverse datasets without architectural changes, demonstrating robust generalization capabilities.
}
\label{fig:cross-dataset}
\end{figure*}

\subsection{Preservation of Pre-trained Knowledge} 
\label{Preservation_Knowledge}
Figure \ref{fig:start}(a) illustrates a performance comparison of Forecast-MAE (FT), Forecast-PEFT (A), and Forecast-PEFT on lane and trajectory reconstruction (pre-training task) and trajectory prediction (downstream task). Forecast-PEFT (A) maintains Forecast-MAE's reconstruction performance (no forgetting) with only 17\% of parameters adjusted, achieving a trajectory prediction minFDE of 1.373, demonstrating its capability to preserve pre-training knowledge while ensuring prediction accuracy. Forecast-PEFT improves prediction with a minFDE of 1.351 using 20\% adapted parameters and only shows a 1.326 increase in reconstruction forgetting. The retained pre-trained decoder in PEFT modules helps avoid catastrophic forgetting, preserving encoder performance and effectively adapting to prediction tasks.

Contrastingly, Forecast-MAE (FT) fully tunes all parameters but does not surpass Forecast-PEFT (A) in performance, while the reconstruction error increases by factor 3, i.e., forgets to reconstruct the trajectories and lanes of pre-training. This indicates that full fine-tuning compromises pre-trained knowledge without enhancing downstream effectiveness, suggesting a potential loss of valuable pre-trained representations during decoder re-learning.

\begin{table*}[t!]
\caption{\textbf{Cross-dataset fine-tuning} results on the AV2, AV1 and nuScenes validation sets, with six prediction modes. Models are pre-trained on AV2 using a masking and reconstruction strategy. ``FT on" and ``Eval on" denote the datasets used for fine-tuning and evaluation, respectively. Forecast-MAE fully fine-tunes all 1.9M parameters for each dataset, while our Forecast-PEFT fine-tunes only the specified modules, totaling 383K parameters. 
}
\label{tab:cross_dataset}
\tablestyle{8pt}{1.1}
\centering
    \resizebox{1.0\linewidth}{!}{
\Huge
\begin{tabular}{cccccccccccccc}
\toprule
& \multicolumn{1}{c}{} & \multicolumn{3}{c}{Eval on AV2 (6s future)} & \multicolumn{3}{c}{Eval on AV2 (3s future)} & \multicolumn{3}{c}{Eval on AV1} & \multicolumn{3}{c}{Eval on nu} \\
\midrule
Method (FT on) & Params & minADE & minFDE & MR & minADE & minFDE & MR & minADE & minFDE & MR & minADE & minFDE & MR \\ \midrule
Forecast-MAE (AV1) & 1.9M & 14.7 & 38.8& 0.967 & 0.76 & 1.69  & 0.248 & \textbf{0.64} & 1.07  & 0.098 & - & -  & - \\
Forecast-MAE (AV2) & 1.9M & 0.72 & 1.43 & 0.183 & \textbf{0.29} & \textbf{0.51}  & 0.026 & 1.75 & 3.42  & 0.406 & 1.31 & 2.80  & 0.450 \\
Forecast-MAE (nu) & 1.9M & 1.92 & 4.26 & 0.557 & - & - & - & - & -  & - & 1.02 & 2.32  & 0.336 \\
\midrule
\baseline{Forecast-PEFT (AV2/AV1/nu)} & \baseline{\textbf{383K/383K/383K}} & \baseline{\textbf{0.71}} & \baseline{\textbf{1.39}} & \baseline{\textbf{0.181}} & \baseline{\textbf{0.29}} & \baseline{\textbf{0.51}} & \baseline{\textbf{0.024}} & \baseline{\textbf{0.64}} & \baseline{\textbf{1.06}} & \baseline{\textbf{0.096}} & \baseline{\textbf{0.99}} & \baseline{\textbf{2.20}} & \baseline{\textbf{0.313}} \\ 
\bottomrule
\end{tabular}
}
\end{table*}

\begin{figure*}[t]
\centering
\begin{minipage}{0.75\linewidth}
\centering
\captionof{table}{\revise{\textbf{Cross-dataset fine-tuning} results on the AV2 and AV1 validation sets, with six prediction modes. Models are pre-trained on AV2 \& AV1 together, using a masking and reconstruction strategy. ``FT on" and ``Eval on" denote the datasets used for fine-tuning and evaluation, respectively. Forecast-MAE fully fine-tunes all 1.9M parameters for each dataset, while our Forecast-PEFT fine-tunes only the specified modules, totaling 383K parameters.}} 
\label{tab:cross_dataset_av2av1}
\tablestyle{8pt}{1.1}
\resizebox{\linewidth}{!}{
\Huge
\begin{tabular}{ccccccccccc}
\toprule
& \multicolumn{1}{c}{} & \multicolumn{3}{c}{Eval on AV2 (6s future)} & \multicolumn{3}{c}{Eval on AV2 (3s future)} & \multicolumn{3}{c}{Eval on AV1} \\
\midrule
Method (FT on) & Params & minADE & minFDE & MR & minADE & minFDE & MR & minADE & minFDE & MR \\ \midrule
Forecast-MAE (AV1) & 1.9M & 14.8 & 39.2 & 0.968 & 0.54 & 1.09  & 0.123 & 0.63 & 1.05  & 0.093 \\
Forecast-MAE (AV2) & 1.9M & 0.72 & 1.43 & 0.183 & \textbf{0.29} & 0.51  & 0.025 & 1.07 & 1.96  & 0.242 \\
\midrule
\baseline{Forecast-PEFT (AV2/AV1)} & \baseline{\textbf{383K/383K}} & \baseline{\textbf{0.70}} & \baseline{\textbf{1.37}} & \baseline{\textbf{0.176}} & \baseline{\textbf{0.29}} & \baseline{\textbf{0.50}} & \baseline{\textbf{0.023}} & \baseline{\textbf{0.61}} & \baseline{\textbf{0.98}} & \baseline{\textbf{0.085}}  \\ 
\bottomrule
\end{tabular}
}
\end{minipage}%
\hfill
\begin{minipage}{0.23\linewidth}
\centering
\includegraphics[width=0.95\linewidth]{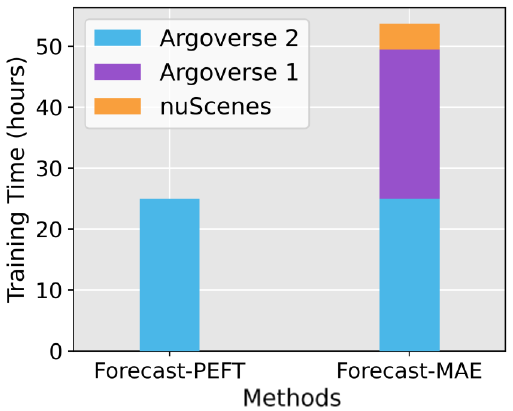}
\caption{\textbf{Pre-training time comparison.} Training time is measured using 4 GPUs.}
\label{fig:training_time}
\end{minipage}
\end{figure*}

\subsection{Cross-Dataset Finetuning}
\label{Cross_Dataset}
Trajectory prediction models often face significant challenges due to distribution shifts when applied across different datasets. Factors such as differing prediction horizons, urban environments, data collection protocols, and object types contribute to these shifts, as observed between the Argoverse 1 and Argoverse 2 datasets \cite{wilson2023argoverse}. These domain gaps can lead to performance degradation, known as catastrophic forgetting, when a model fine-tuned on one dataset is applied to another.

\subsubsection{Data Standardization}
To enable effective cross-dataset fine-tuning, it is crucial to standardize the data from different datasets. This ensures that the model can handle variations in data characteristics, such as sampling rates and prediction horizons, without requiring extensive architectural modifications. In this work, we focus on standardizing data from the Argoverse 1 (AV1), Argoverse 2 (AV2), and nuScenes datasets. \revise{We adopt a unified 10 Hz temporal grid with 5 seconds of history and 6 seconds of prediction. Missing steps are represented as masked entries (implemented as zero padding with validity masks), which are ignored during training and do not affect the loss.}

For the Argoverse datasets, AV1 and AV2, we align the observation and prediction horizons to AV2's longer durations of 5 seconds for observation and 6 seconds for prediction. This contrasts with AV1's shorter horizons of 2 seconds for observation and 3 seconds for prediction. 
\revise{During fine-tuning on AV1, we prepend masked steps to extend the history to 5 seconds and append masked steps to extend the prediction horizon to 6 seconds, as illustrated in Figure~\ref{fig:cross-dataset}.}
 
The nuScenes dataset presents additional challenges due to its different sampling rate of 2Hz compared to AV2's 10Hz. \revise{To ensure consistency, we standardize both datasets to a 10Hz sampling rate by placing the 2 Hz observed points on a 10 Hz grid and representing missing timestamps as masked entries, and then align their historical data durations.} Specifically, we extend nuScenes' original 2-second history to match AV2's 5-second history. Both datasets have the same 6-second prediction horizon. 
Figure \ref{fig:cross-dataset} demonstrates how we address discrepancies in historical data and sampling frequency through strategic padding during the fine-tuning process. This approach ensures that the Forecast-PEFT framework can be efficiently fine-tuned across different datasets without necessitating any changes to its architecture or modules, illustrating its robust flexibility.

\subsubsection{Modular Fine-Tuning Strategy}
Our Forecast-PEFT framework offers a modular and efficient fine-tuning strategy to address the challenges posed by cross-dataset variations. As illustrated in Figure \ref{fig:start}(c), the tunable parameters in Forecast-PEFT can be viewed as a plug-in module. This design allows us to fine-tune only the plug-in module, containing 383K parameters, for each dataset while keeping the rest of the model frozen. This design significantly reduces the number of parameters that need adjustment compared to the comprehensive fine-tuning required by Forecast-MAE, which involves 1.9M parameters per dataset.

The flexibility and robustness of the Forecast-PEFT framework are validated through experiments involving the AV2, AV1, and nuScenes datasets. After pre-training on the AV2 dataset, the model is fine-tuned on AV1 and nuScenes by adjusting only the 383K parameters in the plug-in module. Despite the significant differences among these datasets, our approach achieves results comparable to Forecast-MAE, which requires full fine-tuning of 1.9M parameters for each dataset, as demonstrated in Table \ref{tab:cross_dataset}. This efficiency underscores Forecast-PEFT's capability for effective trajectory prediction across diverse datasets, making it a powerful and flexible tool for this task.

\begin{figure}[bt!]
  \centering
  \includegraphics[width=1.01\linewidth]{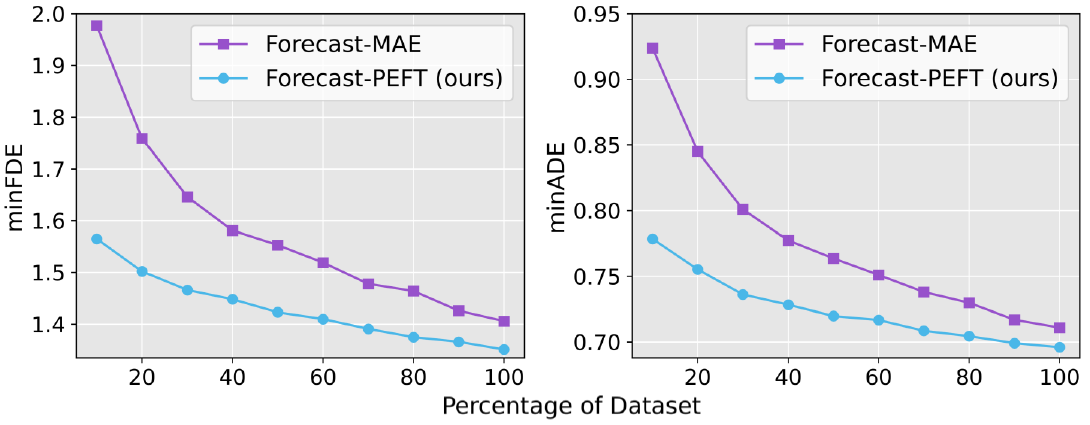}
   \caption{\textbf{Efficiency of using different percentages of the dataset for fine-tuning.} All evaluation results are obtained on the AV2 validation set. }
   \label{fig:rebuttal_startpage}
\end{figure}

\revise{To further extend this analysis, we also evaluate models pre-trained jointly on AV2 and AV1. As shown in Table \ref{tab:cross_dataset_av2av1}, this broader pre-training setup leads to improved cross-dataset generalization compared with AV2-only pre-training (Table \ref{tab:cross_dataset}). In particular, Forecast-PEFT fine-tuned on AV1 achieves better performance on AV1 (minADE: 0.61 vs.\ 0.64, minFDE: 0.98 vs.\ 1.06) compared to pre-training only on AV2, while Forecast-PEFT fine-tuned on AV2 also sees similar improvements, reinforcing the benefit of multi-dataset pre-training while maintaining parameter efficiency.}

In summary, our cross-dataset experiments validate the effectiveness of Forecast-PEFT in maintaining high performance with significantly fewer trainable parameters. This showcases its practical applicability in real-world autonomous driving scenarios where models must generalize across varied environments. The ability to efficiently fine-tune across datasets without requiring extensive architectural changes highlights Forecast-PEFT's potential for broad deployment in diverse autonomous driving contexts.

\begin{figure*}[hbt!]
\begin{center}
\includegraphics[width=1.0\linewidth]{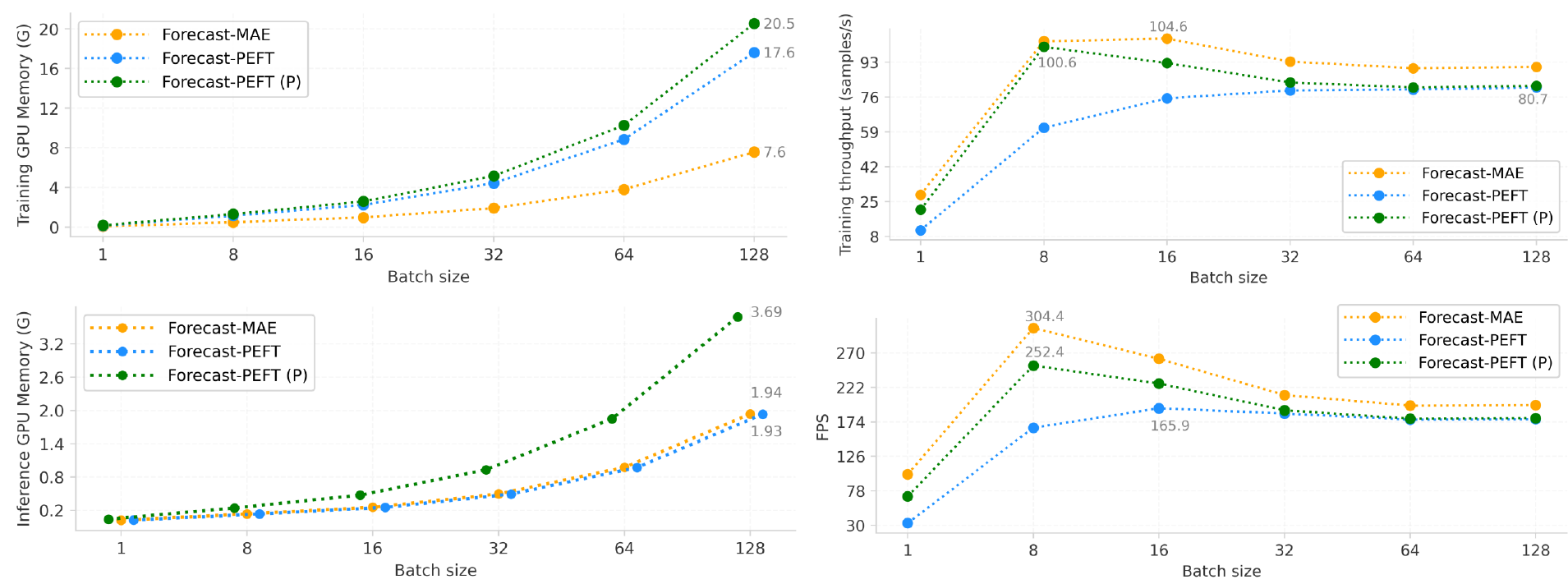}
\end{center}
   \caption{\revise{\textbf{Efficiency analysis comparing Forecast-PEFT, Forecast-PEFT (P), and Forecast-MAE across different batch sizes:} (a) training GPU memory usage, (b) training throughput (samples per second), (c) inference GPU memory usage, and (d) inference speed measured in frames per second (FPS).}
}
\label{fig:hardware_metrics}
\end{figure*}

\subsection{Efficiency Analysis}
\label{Efficiency_Analysis}

Our framework, Forecast-PEFT, is designed with the intention of saving on three critical resources: training time, training data, and storage space. This section outlines the specific advantages of Forecast-PEFT in these areas.

\subsubsection{Training Time} Forecast-PEFT significantly reduces overall training time. As shown in Figure \ref{fig:training_time}, by eliminating the need for pre-training on each dataset, Forecast-PEFT saves 54\% of pre-training time compared to Forecast-MAE. \revise{In terms of absolute training time, pre-training on Argoverse 2, Argoverse 1, and nuScenes takes 
25.0, 24.5, and 4.3 hours, respectively. Consequently, Forecast-MAE requires a total of 53.8 hours 
for pre-training, while Forecast-PEFT only requires 25.0 hours.} Additionally, the fine-tuning time for Forecast-PEFT is comparable to that of Forecast-MAE. For instance, fine-tuning on the Argoverse 2 dataset takes around 5 hours for both Forecast-PEFT and Forecast-MAE.

\subsubsection{Training Data} Forecast-PEFT is also efficient in terms of the amount of training data required. In our evaluation, we investigate the impact of varying the fine-tuning dataset size on the performance of Forecast-PEFT and Forecast-MAE models. This experiment aims to understand how different proportions of the dataset, ranging from 10\% to 100\%, influence the models' predictive accuracy in motion forecasting tasks.
By incrementally increasing the dataset fraction used for fine-tuning, we seek to determine how dataset size limitations affect the performance of each model and whether smaller datasets hinder the models' ability to make accurate predictions.

The results, depicted in Figure \ref{fig:rebuttal_startpage}, show experiments conducted on the AV2 validation set and reveal a consistent trend across two metrics: minADE and minFDE. As the dataset size increases from 10\% to 100\%, both models exhibit a decrease in prediction error, indicating improved forecasting accuracy.

However, the extent of improvement differs between the two models. As shown in Figure \ref{fig:rebuttal_startpage}, Forecast-PEFT can match the performance of Forecast-MAE while using 30\% to 40\% less training data, specifically 30\% less for minADE and 40\% less for minFDE. This underscores the efficiency of Forecast-PEFT in maintaining relatively good performance even with limited data.

Importantly, our Forecast-PEFT model consistently outperforms the Forecast-MAE model across all dataset sizes. This consistent superiority demonstrates the effectiveness of the Forecast-PEFT framework in leveraging limited data, showcasing its potential for reliable performance even when the available data for fine-tuning is constrained.

\subsubsection{Storage Space} In terms of storage space, Forecast-PEFT requires $2.42M + 0.38M \times N$ parameters, while Forecast-MAE requires $1.9M \times N$ parameters (where “$N$” is the number of datasets/domains and “$M$” denotes 1 million parameters). For each new dataset, Forecast-PEFT needs an additional 0.38M parameters compared to 1.9M for the baseline, resulting in significant storage savings as the number of datasets increases.

By leveraging these efficiencies, Forecast-PEFT stands out as a highly effective and practical framework for trajectory prediction across diverse datasets, making it suitable for real-world autonomous driving applications.

\revise{\subsection{Other hardware considerations}
As shown in Figure \ref{fig:hardware_metrics}, we also evaluate the following hardware considerations: training GPU memory usage, training throughput (samples per second), inference GPU memory usage, and inference speed measured in frames per second (FPS).
We evaluate two implementations of our proposed method: Forecast-PEFT and Forecast-PEFT (P), which differ in how the decoding process is handled.
The standard Forecast-PEFT model processes multiple modalities sequentially, running the decoder blocks in a loop for each modality. This iterative approach generates one modality at a time, which conserves memory but requires more computation time.
Forecast-PEFT (P) employs a parallel processing strategy that handles all modalities simultaneously in a single forward pass. By batching all modality inputs together and processing them at once through the decoder, it achieves significantly faster inference speed at the cost of higher memory consumption.
The primary trade-off between these two variants is computational efficiency (Forecast-PEFT (P)) versus memory efficiency (Forecast-PEFT). All these hardware evaluations are conducted on a single NVIDIA A100 GPU. \revise{Specifically, Forecast-PEFT shows slightly lower inference GPU memory usage than Forecast-MAE, 
while Forecast-PEFT (P) achieves inference FPS and training throughput close to those of 
Forecast-MAE. However, both Forecast-PEFT and Forecast-PEFT (P) require higher GPU memory 
during training due to the additional prompt and adapter parameters.}

}

\begin{figure*}[hbt!]
\begin{center}
\includegraphics[width=1.01\linewidth]{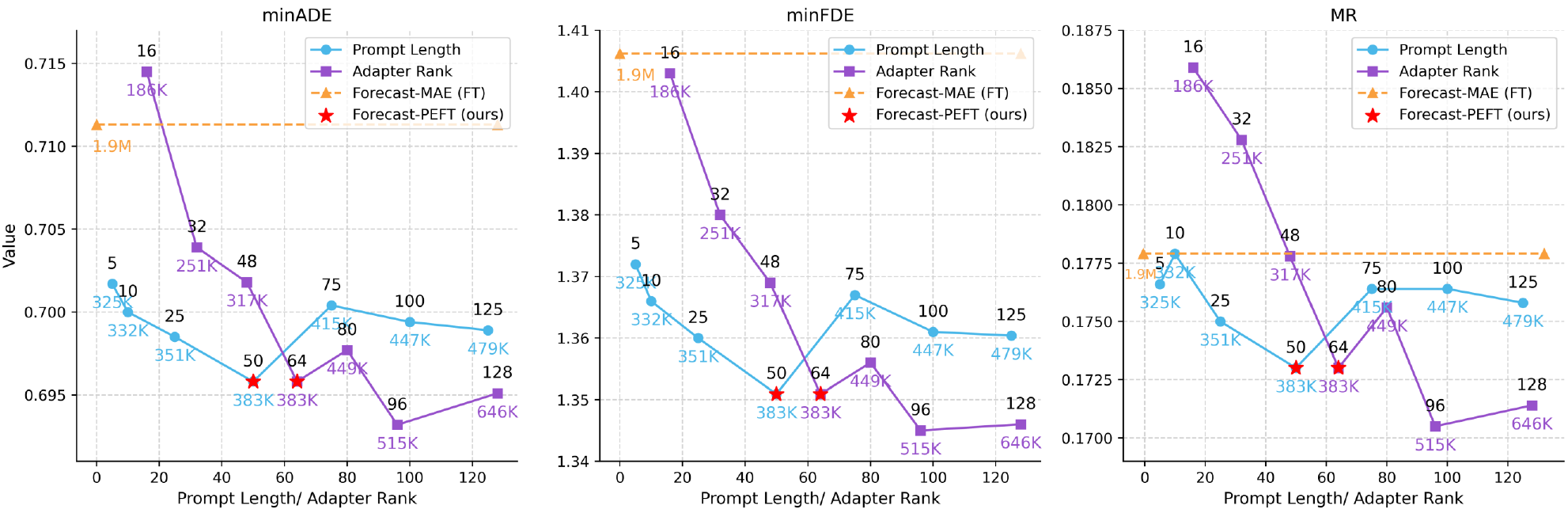}
\end{center}
   \caption{
   \textbf{Ablation study on Prompt Length and Adapter Rank.} 
In our ablation study, we fixed the adapter rank at 64 when assessing prompt length and maintained a constant prompt length of 50 for evaluating adapter rank.
   }
\label{fig:ablation}
\end{figure*}

\begin{table}[!tbp]
\caption{
\textbf{Ablation study for different modules.}  The top gray row represents our Forecast-PEFT (A) model, tuning only the additive modules, while the bottom gray row shows our complete Forecast-PEFT, tuning both additive and selective modules. 
}
\label{tab:components}
\tablestyle{8pt}{1.1}
\resizebox{1.0\linewidth}{!}{
\fontsize{16pt}{14pt} \selectfont
\begin{tabular}{ccccccc}
\toprule
\begin{tabular}[c]{@{}c@{}}Prompt\end{tabular} & \begin{tabular}[c]{@{}c@{}}Head \end{tabular}  & \begin{tabular}[c]{@{}c@{}}LN\&Bias \end{tabular} & \begin{tabular}[c]{@{}c@{}}Adapter\end{tabular} & minADE & minFDE & MR \\ \midrule
 \checkmark & &  &  & 0.775 & 1.56 & 0.223 \\
\checkmark & \checkmark & \multicolumn{1}{l}{} &  & 0.749 & 1.49 & 0.209 \\
\checkmark &  & \checkmark &  & 0.753 & 1.50 & 0.205 \\
\baseline{\checkmark} & \baseline{} & \baseline{} & \baseline{\checkmark} & \baseline{0.704} & \baseline{1.37} & \baseline{0.181} \\
\checkmark & \checkmark & \checkmark &  & 0.735 & 1.45 & 0.197 \\
\checkmark & \checkmark &  & \checkmark & 0.701 & 1.37 & 0.180 \\
\checkmark &  & \checkmark & \checkmark & 0.705 & 1.38 & 0.181 \\
\baseline{\checkmark} & \baseline{\checkmark} & \baseline{\checkmark} & \multicolumn{1}{c}{\baseline{\checkmark}} & \baseline{\textbf{0.696}} & \baseline{\textbf{1.35}} & \baseline{\textbf{0.173}} \\ \midrule
\multicolumn{4}{c}{Forecast-MAE} & 0.711 & 1.41 & 0.178 \\ \bottomrule
\end{tabular}
}
\vspace{1.0em}

\caption{\textbf{Ablation study for different CEP depth.}}
\label{tab:encoder_depth}
\tablestyle{8pt}{1.1}
\resizebox{1.0\linewidth}{!}{
\begin{tabular}{ccccc}
\toprule
CEP depth& Params & minADE & minFDE & MR \\ \midrule
0 & 357K & 0.7027 & 1.368 & 0.1759 \\
1 & 364K & 0.6983 & 1.367 & 0.1764 \\
2 & 370K & 0.6997 & 1.367 & 0.1751 \\
3 & 377K & 0.6981 & 1.361 & 0.1754 \\
\baseline{4} & \baseline{383K} & \baseline{0.6958} & \baseline{1.351} & \baseline{0.1730} \\ \bottomrule
\end{tabular}
}
\end{table}

\subsection{Ablation Study}
\label{Ablation_Study}
\subsubsection{Effectiveness of Main Components}
Table \ref{tab:components} presents an ablation study on the components of Forecast-PEFT using the AV2 validation set. The bottom row of the table shows the baseline results of Forecast-MAE, while the gray rows represent the performance of Forecast-PEFT (A) and the full Forecast-PEFT. Forecast-PEFT (A) exhibits some performance drops in minADE and minFDE compared to complete fine-tuning. However, by selectively unfreezing modules in Forecast-PEFT, the framework outperforms full fine-tuning across all metrics, including MR. This validates the effectiveness of our designed components in the Forecast-PEFT framework.

\subsubsection{Effects of Prompt Length and Depth}
Our ablation studies examine the impact of prompt length and depth for Contextual Embedding Prompts (CEP) and Modality-Control Prompts (MCP), crucial components of the Forecast-PEFT framework. The results offer valuable insights into optimizing these parameters for improved performance.

\begin{figure*}[hbt!]
\begin{center}
\includegraphics[width=1.0\linewidth]{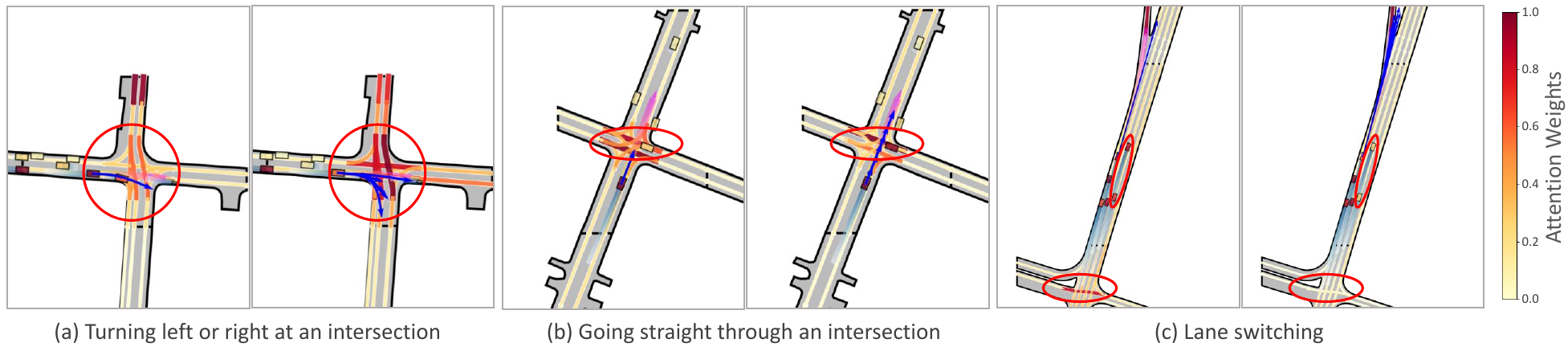}
\end{center}
   \caption{\revise{\textbf{Visualization of attention weights on road agents' history and lane segments}, illustrating the effectiveness of Contextual Embedding Prompts (CEPs). Each scenario represents typical driving maneuvers: (a) turning left or right at intersections, (b) proceeding straight through intersections, and (c) lane switching. 
   \revise{For each case, the left column shows attention maps without CEP, while the right column shows attention maps with CEP, highlighting how CEPs modify the model’s attention toward spatially and behaviorally relevant agents and lane geometry.}} 
}
\label{fig:cep_vis}
\end{figure*}

\textbf{Contextual Embedding Prompt (CEP):}
During \emph{pre-training}, the encoder processes masked historical and future trajectories together with contextual data, whereas during \emph{fine-tuning} and inference it receives only historical trajectories and context. CEPs bridge this input mismatch by directing the encoder’s attention to the most relevant trajectory cues and helping preserve pre-trained knowledge while adapting to new datasets. \revise{Figure~\ref{fig:cep_vis} illustrates how CEPs modify the encoder's attention patterns in typical driving scenarios. 
When the target agent performs a turn, the encoder shifts its attention toward the intersection geometry and the corresponding turning paths. 
In the straight-driving intersection case, the model increases its attention on the ego vehicle's forward lane as well as on approaching vehicles from the right that are also moving straight, together with their corresponding lane geometry. 
In lane-switching scenarios, the encoder shifts attention from broadly covering nearby agents to prioritizing the target lane geometry and the few interacting vehicles that may constrain the lane change, while reducing attention to agents that are unlikely to affect the maneuver. 
Compared to the pre-trained model without PEFT modules, CEPs guide the model to focus more selectively on spatially and behaviorally relevant agents and road structures that are critical for motion forecasting.} 

Our experiments in Table~\ref{tab:encoder_depth} show that increasing CEP depth—i.e., inserting CEPs into more encoder layers—consistently improves performance. Specifically, using CEPs in up to four encoder layers yields the best results, \revise{preserving pre-trained knowledge and keeping the encoder focused on pertinent historical and contextual features}.

\textbf{Modality-Control Prompt (MCP):}
The role of the decoder shifts from reconstructing masked data during pretraining to forecasting future trajectories during fine-tuning. 
\revise{MCPs facilitate this transition by serving as mode-specific queries for multi-modal trajectory generation.}

\revise{We simultaneously vary the prompt lengths of CEP and MCP, ranging from 5 to 125 tokens.} Our findings, depicted in Figure \ref{fig:ablation}, show that longer prompts do not necessarily improve performance, with the optimal length determined to be 50 tokens. 

\subsubsection{Effects of Adapter Rank}
We also explore the effects of adapter rank within the Parallel Adapter (PA) component, drawing inspiration from the LoRA method \cite{hu2021lora}. By varying the adapter's rank from 16 to 128, we aim to balance performance gains against the increase in tunable parameters. Figure \ref{fig:ablation} illustrates that increasing the rank from 16 to 96 enhances performance. However, considering the trade-off with the number of tunable parameters, we select a rank of 64 as the standard, even though a rank of 96 achieves slightly lower errors and missing rate.

\revise{
\subsubsection{Prediction Diversity}
To directly evaluate the separation among prediction modes, we report the Average Pairwise Distance (APD) and Final Pairwise Distance (FPD) on the AV2 validation set. For each sample, APD is computed by averaging the Euclidean distance between every pair of the six predicted trajectories over all future timesteps, while FPD measures the corresponding pairwise distance at the final timestep. Both metrics are then averaged over the validation set.

As shown in Table~\ref{tab:diversity}, Forecast-PEFT achieves an APD of 3.053~m and an FPD of 7.968~m, compared with 2.774~m and 7.158~m for the fully fine-tuned Forecast-MAE baseline. This corresponds to increases of 10.1\% and 11.3\%, respectively. Together with the improved minADE, minFDE, and Miss Rate reported in Table~III, these results show that Forecast-PEFT improves forecasting accuracy without reducing inter-mode separation relative to the fully fine-tuned baseline.}

\revise{These analyses demonstrate the contributions of the individual PEFT components and the effects of prompt length, prompt depth, and adapter rank on the performance of Forecast-PEFT. The prediction diversity results further show that the proposed adaptation does not reduce inter-mode separation relative to the fully fine-tuned baseline. Together, these findings support the design choices of Forecast-PEFT and its effectiveness in achieving accurate multi-modal predictions with limited trainable parameters.}

\begin{table}[h]
\centering
\caption{\revise{\textbf{Inter-mode diversity on the AV2 validation set.} APD and FPD measure pairwise trajectory separation over all timesteps and at the final timestep, respectively.}}
\label{tab:diversity}
\tablestyle{8pt}{1.1}
\begin{tabular}{lcc}
\toprule
\revise{Method} & \revise{APD (m) $\uparrow$} & \revise{FPD (m) $\uparrow$} \\
\midrule
\revise{Forecast-MAE \cite{cheng2023forecast}} & \revise{2.774} & \revise{7.158} \\
\revise{\baseline{Forecast-PEFT}}     & \revise{\baseline{\textbf{3.053}}} & \revise{\baseline{\textbf{7.968}}} \\
\bottomrule
\end{tabular}
\end{table}

\revise{\subsection{Generalization Beyond Forecast-MAE}
To further demonstrate the generality of our PEFT framework, we extend our experiments to Wayformer \cite{nayakanti2023wayformer}, a state-of-the-art transformer-based motion forecasting model with an architecture distinct from Forecast-MAE. In our experiments, we use a public implementation \cite{feng2024unitraj}. We design a masked pre-training strategy for Wayformer, followed by fine-tuning with our PEFT modules. This setup provides a fair comparison with our main experiments and shows that Forecast-PEFT can be readily applied to models beyond Forecast-MAE.

\subsubsection{Masked Pre-training for Wayformer}
We adapt Wayformer \cite{nayakanti2023wayformer} to a masked pre-training pipeline with two main extensions.
First, we expand the input from historical trajectories only to include both history and future trajectories, together with lane polylines.
Second, we repurpose the decoder queries: instead of directly generating multi-modal futures as in the original Wayformer, they are allocated to three groups responsible for reconstructing past trajectories, future trajectories, and lane polylines.
For each scene, we mask half of the agents in the past and the complementary half in the future, and also mask half of the lane polylines.
The encoder processes the masked trajectories and polylines with masking indicators, and the decoder reconstructs only the missing elements.
This pre-training stage teaches Wayformer universal trajectory–map representations aligned with our Forecast-PEFT framework. 

\begin{table}[tb!]
\caption{\revise{\textbf{Quantitative Results on the AV2 Validation set.}  
}}
\label{tab:leaderboard_wayformer}
\centering
    \resizebox{1.0\linewidth}{!}{
  \fontsize{22pt}{18pt} \selectfont
  \begin{tabular}{lllll}
    \toprule
    \textbf{Method} & \textbf{minADE ($\downarrow$)} & \textbf{minFDE ($\downarrow$)} & \textbf{MR ($\downarrow$)} & \textbf{b-minFDE} ($\downarrow$) \\ 
    \midrule
    \revise{AutoBots~\cite{girgis2021latent}} & \hspace{0.4cm}0.85 & \hspace{0.3cm}1.70 & \hspace{0.1cm}0.27 & \hspace{0.5cm}2.51 \\
    \revise{MTR~\cite{shi2022motion}} & \hspace{0.4cm}0.85 & \hspace{0.3cm}1.68 & \hspace{0.1cm}0.30 & \hspace{0.5cm}2.08 \\
    \revise{Wayformer~\cite{nayakanti2023wayformer}} & \hspace{0.4cm}0.85 & \hspace{0.3cm}1.75 & \hspace{0.1cm}0.28 & \hspace{0.5cm}2.38 \\
    \midrule
    Wayformer (w.pretrain)  & \hspace{0.4cm}0.84 & \hspace{0.3cm}1.71 & \hspace{0.1cm}0.27 & \hspace{0.5cm}2.33 \\
    \baseline{Forecast-PEFT (ours)} & \hspace{0.4cm}\baseline{0.83} & \hspace{0.3cm}\baseline{1.69} & \hspace{0.1cm}\baseline{0.27} & \hspace{0.5cm}\baseline{2.30} \\ 
    \baseline{Forecast-FT (ours)} & \hspace{0.4cm}\baseline{0.77 } & \hspace{0.3cm}\baseline{1.51} & \hspace{0.1cm}\baseline{0.22 } & \hspace{0.5cm}\baseline{2.13 } \\
    \bottomrule
    \end{tabular}
     }
\end{table}

\begin{table}[tb!]
\caption{\revise{\textbf{Cross-dataset fine-tuning} results on the AV2 and Nu validation sets, with six prediction modes. Models are pre-trained on AV2 using a masking and reconstruction strategy.} 
}
\label{tab:cross_dataset_wayformer}
\tablestyle{8pt}{1.1}
\centering
    \resizebox{1.0\linewidth}{!}{
\fontsize{30pt}{26pt} \selectfont
\begin{tabular}{ccccccc}
\toprule
& \multicolumn{1}{c}{} & \multicolumn{1}{c}{Eval on AV2} &  \multicolumn{4}{c}{Eval on Nu} \\
\midrule
Method (FT on) & minADE & minFDE & MR & minADE & minFDE & MR \\ \midrule

Wayformer (AV2) & 0.84 & 1.71 & 0.27 & 1.25 & 2.64  & 0.41 \\
Wayformer (Nu)  & 1.72 & 3.71 & 0.56 & 0.98 & 2.25  & 0.38 \\

\midrule

\baseline{Forecast-PEFT (AV2/Nu)} & \baseline{\textbf{0.83}} & \baseline{\textbf{1.69}} & \baseline{\textbf{0.27}} &  \baseline{\textbf{0.98}} & \baseline{\textbf{2.17}} & \baseline{\textbf{0.34}}  \\ 
\bottomrule
\end{tabular}
}
\end{table}

\subsubsection{PEFT Adaptation of Wayformer}
After pre-training, we fine-tune Wayformer using our Forecast-PEFT modules. Specifically, CEPs are added at the encoder input to align masked pre-training inputs with the complete trajectory–map data used during fine-tuning. MCPs are injected into the decoder queries to transition its role from reconstruction to multi-modal forecasting. PAs are integrated into each transformer block of both encoder and decoder to refine local trajectory features without changing backbone parameters. This targeted insertion preserves the strengths of Wayformer’s architecture while enabling parameter-efficient adaptation.

\subsubsection{Results on Wayformer}
We report results on Wayformer \cite{nayakanti2023wayformer} to assess the generalization of Forecast-PEFT.
For consistency with the original design, we use 2.1 seconds of historical input instead of 5 seconds.
The efficiency advantage of PEFT also holds here: Wayformer has 15.2M parameters in total, and our modules contribute only about 15\%, consistent with the ratio observed in earlier experiments.
As shown in Table~\ref{tab:leaderboard_wayformer} and Table~\ref{tab:cross_dataset_wayformer}, Forecast-PEFT achieves improvements comparable to those on Forecast-MAE in both single-dataset and cross-dataset settings, confirming that the modules integrate seamlessly into diverse architectures.

}

\subsection{Limitations} 
The efficacy of Forecast-PEFT largely depends on the quality and relevance of the pre-trained models. The success of fine-tuning can greatly fluctuate depending on the alignment between pre-training and the specific needs of motion forecasting tasks. Additionally, the scalability of Forecast-PEFT for larger datasets or more complex forecasting scenarios requires further investigation.

\revise{Moreover, Forecast-PEFT adopts the same winner-takes-all (WTA) objective as the Forecast-MAE fine-tuning baseline. While the pairwise-distance analysis in Section~\ref{Ablation_Study} shows that Forecast-PEFT does not reduce inter-mode separation relative to the fully fine-tuned baseline, these metrics primarily measure geometric separation and do not fully characterize the quality or semantic validity of prediction diversity. Evaluating diversity-aware training objectives and more comprehensive diversity metrics remains an important direction for future work.}

\revise{Finally, our validation is limited to two Transformer-based encoder--decoder architectures evaluated under the masked-reconstruction pre-training and adaptation setting considered in this work. The effectiveness of Forecast-PEFT on diffusion-based trajectory predictors has not yet been validated. Such an extension requires a separate study, as timestep conditioning, iterative denoising, and diffusion-based sampling differ substantially from the direct reconstruction-to-forecasting transition examined here.}

\subsection{Potential Future Research Avenues} 
Our work can serve as a baseline for exploring PEFT methods in motion forecasting. With more datasets and larger models becoming available, the advantages of our method will become more evident. Through our approach, pre-training on one large dataset and then applying PEFT across various datasets can significantly benefit real-world applications.

\section{Conclusion} 
\label{sec:conclusion}
Our research presents Forecast-PEFT, a parameter-efficient fine-tuning framework specifically designed for motion forecasting in autonomous driving. We discovered that traditional full fine-tuning methods are suboptimal, as they do not fully exploit the synergy between self-supervised learning tasks and trajectory prediction. Forecast-PEFT addresses this issue by effectively utilizing the pre-trained model’s encoder and decoder, considerably reducing the number of trainable parameters without sacrificing performance. This efficiency is crucial for deploying advanced autonomous systems where computational resources are limited. The framework's robustness is further demonstrated through successful cross-dataset fine-tuning, highlighting its versatility and broad applicability.

\section{Acknowledgment}
Juergen Gall has been supported by the ERC Consolidator Grant FORHUE (101044724).

\bibliographystyle{splncs04}
\bibliography{main}

\end{document}